\documentclass{article}

\usepackage{arxiv}

\usepackage[utf8]{inputenc} 
\usepackage[T1]{fontenc}    
\usepackage{hyperref}       
\usepackage{url}            
\usepackage{booktabs}       
\usepackage{amsfonts}       
\usepackage{amssymb}        
\usepackage{subcaption}     
\usepackage{nicefrac}       
\usepackage{microtype}      
\usepackage{graphicx}
\usepackage{natbib}
\usepackage{doi}
\usepackage{amsmath}
\usepackage{listings}
\usepackage{xcolor}
\title{MIP Candy: A Modular PyTorch Framework for Medical Image Processing}

\author{ \href{https://orcid.org/0009-0007-2342-5350}{\includegraphics[scale=0.06]{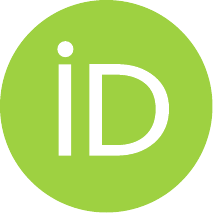}\hspace{1mm}Tianhao Fu}\thanks{Equal contribution.}\\
	University of Toronto, Toronto, ON, Canada\\
	Vector Institute, Toronto, ON, Canada\\
	Project Neura, Toronto, ON, Canada\\
	UTMIST, Toronto, ON, Canada\\
	\texttt{terry.fu@projectneura.org} \\
	\And
	\href{https://orcid.org/0009-0000-9492-8958}{\includegraphics[scale=0.06]{figures/orcid.pdf}\hspace{1mm}Yucheng Chen}\footnotemark[1]\\
	Project Neura, Toronto, ON, Canada\\
	Amplimit, Toronto, ON, Canada\\
	\texttt{steven.chen@projectneura.org} \\
}

\date{}

\renewcommand{\headeright}{Technical Report}
\renewcommand{\undertitle}{Technical Report}
\renewcommand{\shorttitle}{MIP Candy}

\hypersetup{
	pdftitle={MIP Candy: A Modular PyTorch Framework for Medical Image Processing},
	pdfsubject={cs.LG; cs.AI; cs.CV},
	pdfauthor={Tianhao Fu, Yucheng Chen}
}

\begin{document}
	\maketitle
	
	\begin{abstract}
		Medical image processing demands specialized software that handles high-dimensional volumetric data, heterogeneous file formats, and domain-specific training procedures.
		Existing frameworks either provide low-level components that require substantial integration effort or impose rigid, monolithic pipelines that resist modification.
		We present MIP Candy (MIPCandy), a freely available, PyTorch-based framework designed specifically for medical image processing.
		MIPCandy provides a complete, modular pipeline spanning data loading, training, inference, and evaluation, allowing researchers to obtain a fully functional process workflow by implementing a single method---\texttt{build\_network}---while retaining fine-grained control over every component.
		Central to the design is \texttt{LayerT}, a deferred configuration mechanism that enables runtime substitution of convolution, normalization, and activation modules without subclassing.
		The framework further offers built-in $k$-fold cross-validation, dataset inspection with automatic region-of-interest detection, deep supervision, exponential moving average, multi-frontend experiment tracking (Weights \& Biases, Notion, MLflow), training state recovery, and validation score prediction via quotient regression.
		An extensible bundle ecosystem provides pre-built model implementations that follow a consistent trainer--predictor pattern and integrate with the core framework without modification.
		MIPCandy is open-source under the Apache-2.0 license and requires Python~3.12 or later.
		Source code and documentation are available at \url{https://github.com/ProjectNeura/MIPCandy}.
	\end{abstract}
	
	\section{Introduction}
	\label{sec:introduction}
	
	Medical image segmentation---the task of assigning a semantic label to each voxel in a clinical scan---is a fundamental step in computer-aided diagnosis, treatment planning, and longitudinal disease monitoring.
	Unlike natural images, medical data are typically stored in domain-specific formats (NIfTI, DICOM, MHA) that encode acquisition metadata such as voxel spacing, orientation, and modality.
	Volumes are often three-dimensional, high-resolution, and acquired with anisotropic spacing, making both the data handling and the model training substantially more involved than in standard computer vision pipelines.
	Meanwhile, expert annotations are scarce and expensive, placing a premium on training strategies---cross-validation, data augmentation, deep supervision---that extract maximal information from limited labels.
	
	General-purpose deep learning frameworks such as PyTorch~\citep{pytorch} and TensorFlow~\citep{tensorflow} provide the computational substrate but lack medical-imaging-specific functionality.
	Building a segmentation pipeline from scratch on top of these frameworks requires implementing format-aware data loading, geometry-preserving transforms, specialized loss functions, sliding window inference for large volumes, and reproducible experiment management---a significant engineering effort that is duplicated across research groups.
	
	Several domain-specific frameworks have been developed to address this gap, ranging from comprehensive component libraries such as MONAI~\citep{monai} and TorchIO~\citep{torchio} to fully automated pipelines such as nnU-Net~\citep{nnunet}.
	However, existing solutions tend toward one of two extremes: they either provide low-level components that require substantial assembly effort or impose monolithic pipelines that resist modification.
	We review these approaches in Section~\ref{sec:related}.
	
	We present MIP Candy (hereafter MIPCandy), a PyTorch-based framework designed to occupy the middle ground between these extremes.
	MIPCandy provides a \emph{complete} pipeline---from data loading and dataset inspection through training and inference to evaluation---yet every component is independently usable and replaceable.
	A researcher can obtain a fully functional segmentation workflow by implementing a single abstract method, \texttt{build\_network}, on top of the provided \texttt{SegmentationTrainer} preset; alternatively, individual modules such as the metric functions, the visualization utilities, or the dataset classes can be adopted incrementally into an existing PyTorch codebase.
	
	The principal contributions of this work are as follows:
	\begin{enumerate}
		\item \textbf{LayerT}, a deferred module configuration mechanism that enables runtime substitution of convolution, normalization, and activation layers without subclassing (Section~\ref{sec:layert}).
		\item A \textbf{hierarchical training framework} with pre-configured segmentation presets, deep supervision, exponential moving average, training state recovery, and multi-frontend experiment tracking (Section~\ref{sec:training}).
		\item A \textbf{dataset inspection system} that automatically computes foreground bounding boxes, class distributions, and intensity statistics, enabling region-of-interest-based patch sampling (Section~\ref{sec:data}).
		\item \textbf{Validation score prediction} via quotient regression, which fits a rational function to the validation trajectory and estimates both the maximum achievable score and the optimal stopping epoch (Section~\ref{sec:training}).
		\item An extensible \textbf{bundle ecosystem} that packages model architectures, trainers, and predictors into self-contained, reusable units (Section~\ref{sec:bundles}).
	\end{enumerate}
	
	MIPCandy requires Python~3.12 or later and makes deliberate use of modern language features---type aliases (PEP~613), pattern matching, the \texttt{Self} type, and the \texttt{@override} decorator---to improve readability and catch errors at development time.
	The framework is released under the Apache-2.0 license at \url{https://github.com/ProjectNeura/MIPCandy}.

	\section{Related Work}
	\label{sec:related}
	
	Existing software for medical image segmentation can be broadly organized into three categories: general-purpose deep learning frameworks, domain-specific component libraries, and end-to-end segmentation pipelines.
	
	\paragraph{General-purpose frameworks.}
	PyTorch~\citep{pytorch} and TensorFlow~\citep{tensorflow} provide the foundational building blocks---automatic differentiation, GPU-accelerated tensor operations, and modular neural network layers---on which all contemporary medical imaging tools are built.
	Higher-level wrappers such as PyTorch Lightning~\citep{lightning} reduce boilerplate by standardizing the training loop, checkpoint management, and distributed training.
	However, none of these frameworks are aware of the particularities of medical data: volumetric file formats, voxel spacing, anisotropic resolution, or the class-imbalance and small-dataset regimes that are characteristic of clinical annotations.
	Researchers building on these frameworks must therefore implement format-aware data loading, geometry-preserving transforms, specialized loss functions, and reproducible experiment management from scratch---an engineering effort that is duplicated across groups.
	
	\paragraph{Domain-specific component libraries.}
	MONAI~\citep{monai} is the most widely adopted medical imaging library for PyTorch.
	It provides a large collection of transforms (spatial, intensity, crop/pad, with both array and dictionary interfaces), network architectures, loss functions, and metrics, together with Ignite-based training engines.
	MONAI follows an opt-in, compositional design: individual components can be imported independently and composed with vanilla PyTorch code.
	This flexibility, however, comes at the cost of assembly effort.
	Constructing a complete training pipeline in MONAI requires the user to select and configure each component---data loaders, transform chains, network, optimizer, loss, metric, engine, and event handlers---and wire them together manually.
	There is no single entry point that produces a working segmentation workflow with researched defaults.
	
	TorchIO~\citep{torchio} focuses on a narrower scope: efficient loading, preprocessing, augmentation, and patch-based sampling of medical images.
	It integrates well with PyTorch's \texttt{DataLoader} and supports queue-based patch extraction for large 3D volumes.
	TorchIO is complementary to, rather than competitive with, pipeline frameworks; it addresses data handling but does not provide training loops, experiment management, or evaluation utilities.
	
	\paragraph{End-to-end segmentation pipelines.}
	nnU-Net~\citep{nnunet} occupies the opposite end of the spectrum.
	Given a dataset in a prescribed format, it automatically determines the preprocessing strategy, network topology, training schedule, and post-processing, achieving state-of-the-art results on a wide range of benchmarks~\citep{nnunetrevisited}.
	This automation, however, comes at the cost of modularity.
	The pipeline's components---data augmentation, architecture selection, loss function, and training loop---are tightly coupled and not designed to be used independently.
	Substituting a custom network architecture, loss function, or training strategy requires modifying nnU-Net's internal code rather than composing external modules.
	Furthermore, the training process provides limited real-time visibility: intermediate predictions, per-epoch metric trajectories, and estimated time to completion are not surfaced to the user during training.
	MIST~\citep{mist} is a more recent end-to-end framework that similarly automates preprocessing and training for 3D medical image segmentation, though with a simpler, more configurable pipeline than nnU-Net.
	
	\paragraph{Earlier efforts.}
	NiftyNet~\citep{niftynet}, built on TensorFlow, was among the first open-source platforms dedicated to medical image analysis, providing configurable pipelines for segmentation, regression, and image generation.
	DLTK~\citep{dltk} offered reference deep learning implementations for medical imaging, and DeepNeuro~\citep{deepneuro} targeted neuroimaging workflows.
	All three projects are now largely unmaintained and incompatible with current versions of their underlying frameworks.
	
	\paragraph{Positioning.}
	Table~\ref{tab:comparison} summarizes the capabilities of the most relevant active frameworks.
	MIPCandy is designed to combine the completeness of an end-to-end pipeline with the modularity of a component library.
	Like nnU-Net, it provides a fully configured training workflow with researched defaults---a working segmentation pipeline can be obtained by implementing a single method (\texttt{build\_network}).
	Like MONAI, every component is independently usable and replaceable.
	Unlike both, MIPCandy emphasizes \emph{training transparency}: per-epoch metric curves, input--label--prediction previews, validation score prediction with estimated time to completion, and multi-frontend experiment tracking are built into the training loop rather than requiring external configuration.
	
	\begin{table}[t]
		\centering
		\caption{Feature comparison of active medical image segmentation frameworks.}
		\label{tab:comparison}
		\begin{tabular}{@{}lcccc@{}}
			\toprule
			Feature & nnU-Net & MONAI & TorchIO & MIPCandy \\
			\midrule
			Complete training pipeline     & \checkmark & --         & --         & \checkmark \\
			One-method setup               & \checkmark & --         & --         & \checkmark \\
			Modular / individually usable  & --         & \checkmark & \checkmark & \checkmark \\
			Custom architecture swap       & Hard       & Manual     & N/A        & \texttt{build\_network} \\
			Deep supervision               & \checkmark & Manual     & N/A        & One flag \\
			EMA support                    & --         & Manual     & N/A        & One flag \\
			Training state recovery        & \checkmark & Manual     & N/A        & Built-in \\
			Real-time metric visualization & --         & Via handlers & N/A      & Built-in \\
			Prediction previews            & --         & --         & N/A        & Built-in \\
			Score prediction / ETC         & --         & --         & --         & \checkmark \\
			Multi-frontend tracking        & TensorBoard & TensorBoard & N/A     & WandB / Notion / MLflow \\
			Dataset inspection \& ROI      & Internal   & --         & --         & \texttt{inspect()} \\
			Patch-based sampling           & \checkmark & \checkmark & \checkmark & \checkmark \\
			$k$-fold cross-validation      & \checkmark & --         & --         & \checkmark \\
			Bundle / model ecosystem       & --         & MONAI Bundles & --      & \checkmark \\
			\bottomrule
		\end{tabular}
	\end{table}

	\section{Design Philosophy}
	\label{sec:design}

	MIPCandy is guided by four design principles that together shape the framework's API, implementation, and extension model.
	
	\paragraph{PyTorch-native.}
	Every trainable component in MIPCandy is a standard \texttt{nn.Module}; every dataset is a standard \texttt{torch.utils.data.Dataset}.
	Loss functions, normalization layers, padding operators, and deep supervision wrappers are all \texttt{nn.Module} subclasses that compose with the rest of the PyTorch ecosystem without adaptation.
	As a consequence, any existing PyTorch utility---distributed data parallelism, automatic mixed precision, \texttt{torch.compile}---can be applied to MIPCandy components without modification.
	
	\paragraph{Opt-in and incremental.}
	No module assumes that the rest of the framework is present.
	A researcher can adopt a single component---a loss function, a dataset class, a metric---into an existing codebase and later integrate additional modules as needed.
	
	\paragraph{Composition over inheritance.}
	MIPCandy favors runtime configuration over class proliferation.
	The \texttt{LayerT} mechanism (Section~\ref{sec:layert}) stores a module type together with its constructor arguments and instantiates the module on demand, enabling users to swap convolution, normalization, or activation layers by passing different \texttt{LayerT} instances rather than defining new subclasses.
	The same compositional approach appears throughout: \texttt{DeepSupervisionWrapper} wraps any loss module, \texttt{BinarizedDataset} wraps any supervised dataset, and \texttt{TrainerToolbox} bundles model, optimizer, scheduler, and criterion into a flat dataclass.
	
	\paragraph{Minimal API surface.}
	The common case should require no configuration.
	\texttt{SegmentationTrainer} ships with a pre-configured optimizer, scheduler, and loss that selects the appropriate variant based on the number of classes.
	A complete training run can be launched with \texttt{trainer.train(100)}; all optional keyword arguments have researched defaults.
	Conversely, every default is overridable via class attributes or method overrides.

	\section{System Architecture}
	\label{sec:architecture}

	MIPCandy is organized into nine loosely coupled modules, summarized in Table~\ref{tab:modules}.
	Each module can be imported and used independently; the training framework, for example, has no compile-time dependency on the evaluation module, and the metrics module depends only on PyTorch tensors.
	
	\begin{table}[h]
		\centering
		\caption{MIPCandy module overview.}
		\label{tab:modules}
		\begin{tabular}{@{}lp{9cm}@{}}
			\toprule
			Module & Responsibility \\
			\midrule
			\texttt{mipcandy.data} & Multi-format I/O, dataset classes, $k$-fold cross-validation, transforms, dataset inspection, visualization \\
			\texttt{mipcandy.layer} & \texttt{LayerT} configuration, device management, checkpoint I/O, \texttt{WithNetwork} and \texttt{WithPaddingModule} base classes \\
			\texttt{mipcandy.training} & \texttt{Trainer} base class, \texttt{TrainerToolbox} dataclass, experiment management, validation score prediction \\
			\texttt{mipcandy.presets} & \texttt{SegmentationTrainer} preset with pre-configured loss, optimizer, scheduler, and deep supervision \\
			\texttt{mipcandy.inference} & \texttt{Predictor} base class, \texttt{parse\_predictant} utility, file-level prediction and export \\
			\texttt{mipcandy.evaluation} & \texttt{Evaluator} class, \texttt{EvalResult} container with per-case and aggregate metrics \\
			\texttt{mipcandy.metrics} & Dice-family metrics: \texttt{binary\_dice}, \texttt{dice\_similarity\_coefficient}, \texttt{soft\_dice} \\
			\texttt{mipcandy.frontend} & Experiment tracking frontends: Weights~\&~Biases, Notion, MLflow, and hybrid combinations \\
			\texttt{mipcandy.common} & Building blocks: convolution blocks, loss functions, learning rate schedulers, quotient regression \\
			\bottomrule
		\end{tabular}
	\end{table}
	
	The remainder of this section describes each module in detail.

	\subsection{Data Pipeline}
	\label{sec:data}

	\paragraph{Multi-format I/O.}
	MIPCandy reads and writes medical images via SimpleITK~\citep{simpleitk}, supporting NIfTI, MetaImage, and raster formats.
	The \texttt{load\_image()} function performs automatic format detection, optional isotropic resampling, and direct device placement.
	For intermediate storage, \texttt{fast\_save()} and \texttt{fast\_load()} use the safetensors format~\citep{safetensors}, providing zero-copy deserialization.
	
	\paragraph{Dataset hierarchy.}
	All datasets inherit from a generic base that extends \texttt{torch.utils.data.Dataset} and provides device management, $k$-fold splitting, and a path-export interface.
	Key implementations include \texttt{NNUNetDataset} (nnU-Net raw format with multimodal support), \texttt{BinarizedDataset} (multiclass-to-binary wrapper), and composition utilities for merging datasets.
	Every dataset exposes a \texttt{fold()} method for $k$-fold cross-validation with configurable splitting strategies.
	
	\paragraph{Dataset inspection.}
	The \texttt{inspect()} function scans a supervised dataset and records per-case foreground bounding boxes, class distributions, and intensity statistics.
	From these annotations the framework computes a \emph{statistical foreground shape} and derives a region-of-interest (ROI) shape for patch-based training.
	\texttt{RandomROIDataset} samples random patches with configurable foreground oversampling (default: 33\% of patches contain foreground).

	\subsection{LayerT Configuration System}
	\label{sec:layert}

	Neural network architectures are typically parameterized by the choice of convolution, normalization, and activation layers.
	The standard approaches to making these choices configurable are either to accept many constructor arguments or to require subclassing for each combination.
	Both scale poorly: a network that supports 2D and 3D convolutions, batch and group normalization, and multiple activations would need $2 \times 2 \times k$ subclasses under an inheritance-based approach.
	
	\texttt{LayerT} solves this by storing a module \emph{type} together with its constructor keyword arguments as a lightweight descriptor.
	The module is instantiated only when \texttt{assemble()} is called, at which point positional and keyword arguments are merged with the stored defaults:
	
	\begin{lstlisting}
		from mipcandy.layer import LayerT
		from torch import nn
		
		# Define layer configurations
		conv  = LayerT(nn.Conv2d)
		norm  = LayerT(nn.BatchNorm2d, num_features="in_ch")
		act   = LayerT(nn.ReLU, inplace=True)
		
		# Instantiate at build time
		conv_module = conv.assemble(64, 128, 3, padding=1)  # nn.Conv2d(64, 128, 3, padding=1)
		norm_module = norm.assemble(in_ch=128)               # nn.BatchNorm2d(128)
		act_module  = act.assemble()                          # nn.ReLU(inplace=True)
	\end{lstlisting}
	
	The string \texttt{"in\_ch"} acts as a \emph{deferred parameter}: it is resolved to the integer value passed to \texttt{assemble()}, allowing a single descriptor to adapt to different channel counts.
	MIPCandy uses \texttt{LayerT} pervasively; for example, \texttt{ConvBlock2d} accepts \texttt{LayerT} arguments for convolution, normalization, and activation, with pre-configured defaults that can be overridden at construction time:
	
	\begin{lstlisting}
		from mipcandy.common import ConvBlock2d
		from mipcandy.layer import LayerT
		from torch import nn
		
		# Default: Conv2d + BatchNorm2d + ReLU
		block = ConvBlock2d(64, 128, 3, padding=1)
		
		# Custom: Conv2d + GroupNorm + GELU
		block = ConvBlock2d(
		64, 128, 3, padding=1,
		norm=LayerT(nn.GroupNorm, num_groups=8, num_channels="in_ch"),
		act=LayerT(nn.GELU),
		)
	\end{lstlisting}

	\subsection{Training Framework}
	\label{sec:training}

	\paragraph{Trainer and TrainerToolbox.}
	The \texttt{Trainer} base class manages the training lifecycle.
	Training state is encapsulated in a \texttt{TrainerToolbox} dataclass that bundles the model, optimizer, scheduler, criterion, and an optional EMA~\citep{ema} model.
	The toolbox is constructed from builder methods (\texttt{build\_network}, \texttt{build\_optimizer}, etc.) that subclasses override to customize each component.
	Each training run produces a timestamped experiment folder containing checkpoints, per-epoch metrics (CSV), progress plots, log files, and worst-case prediction previews (see Section~\ref{sec:transparency}).
	Before the first epoch, a sanity check validates the output shape and reports MACs and parameter count.
	Training state is serialized every epoch, enabling seamless recovery after interruptions.
	
	\paragraph{SegmentationTrainer preset.}
	\texttt{SegmentationTrainer} extends \texttt{Trainer} with pre-configured defaults: a combined Dice--cross-entropy loss~\citep{dice} that selects the binary or multiclass variant automatically, SGD with momentum 0.99 and Nesterov acceleration, a polynomial learning rate scheduler, and gradient clipping.
	When the \texttt{deep\_supervision} flag is set, the criterion is wrapped in a \texttt{DeepSupervisionWrapper}~\citep{deepsupervision} with auto-computed weights $w_i = 2^{-i}$.
	EMA via PyTorch's \texttt{AveragedModel} can be enabled with a single flag.
	
	\paragraph{Validation score.}
	MIPCandy defines the validation score as the negated combined loss: $s = -\mathcal{L}_{\text{val}}$.
	This convention maps every loss function to a unified ``higher is better'' scale, so that best-checkpoint selection, early stopping, and score prediction all use a single comparison direction ($s_{\text{new}} > s_{\text{best}}$) regardless of the underlying criterion.
	The framework then fits a quotient regression model---a rational function $P(x)/Q(x)$---to the validation score trajectory, estimating the maximum achievable score and the epoch at which it will be reached (ETC).
	
	\paragraph{Frontend integrations.}
	Experiment tracking uses a pluggable \texttt{Frontend} protocol; shipped implementations cover Weights~\&~Biases~\citep{wandb}, Notion, and MLflow~\citep{mlflow}, with a factory for combining multiple frontends.
	
	The visual aspects of training transparency---console output, metric plots, prediction previews, and frontend screenshots---are presented in Section~\ref{sec:transparency}.

	\subsection{Inference and Evaluation}
	\label{sec:inference}
	
	The \texttt{Predictor} class mirrors the trainer's \texttt{WithNetwork} interface: the user implements \texttt{build\_network()} and the framework handles lazy checkpoint loading, device placement, and padding.
	A unified \texttt{parse\_predictant()} function accepts file paths, directories, tensors, or datasets, normalizing them into a common format.
	Predictors support single-image, batch, and file-level output (\texttt{.png} for 2D, \texttt{.mha} for 3D).
	
	The \texttt{Evaluator} class accepts arbitrary metric functions and produces an \texttt{EvalResult} container with per-case and aggregate scores, supporting evaluation from datasets, raw tensors, or end-to-end predict-and-evaluate workflows.
	MIPCandy provides Dice-family metrics---\texttt{binary\_dice}, \texttt{dice\_similarity\_coefficient}, and \texttt{soft\_dice}---covering boolean, one-hot, and soft-probability formats.
	The same functions serve dual roles as both loss components and evaluation metrics.

	\section{Bundle Ecosystem}
	\label{sec:bundles}

	While the core framework provides the infrastructure for training, inference, and evaluation, specific network architectures and their associated configurations are distributed as \emph{bundles}---self-contained packages that plug into the framework without modifying it.
	
	\paragraph{Bundle structure.}
	Each bundle follows a consistent three-file pattern:
	\begin{itemize}
		\item \textbf{Model}: an \texttt{nn.Module} subclass implementing the architecture, plus builder functions (\texttt{make\_unet2d}, \texttt{make\_unet3d}, etc.) that construct common configurations.
		\item \textbf{Trainer}: a class extending \texttt{SegmentationTrainer} that overrides \texttt{build\_network()} (and optionally \texttt{build\_padding\_module()}, \texttt{build\_optimizer()}, or \texttt{backward()}).
		\item \textbf{Predictor}: a class extending \texttt{Predictor} that overrides \texttt{build\_network()}.
	\end{itemize}
	
	The only mandatory override is \texttt{build\_network()}, which receives the shape of a single input tensor and returns an \texttt{nn.Module}.
	All other training infrastructure---loss, optimizer, scheduler, checkpointing, metric tracking, deep supervision---is inherited from the preset.
	
	\paragraph{Integration.}
	Bundles depend on the core framework through its public API and use \texttt{LayerT}, presets, and data pipeline classes directly.
	No monkey-patching or registration is required.
	Bundle-specific behavior (e.g., custom normalization selection based on batch size, architecture-specific deep supervision) is expressed through standard method overrides.
	
	\paragraph{Extensibility.}
	The bundle mechanism is not limited to model architectures.
	Augmentation pipelines, loss functions, and task-specific workflows can all be packaged as bundles.
	At the time of writing, MIPCandy ships with bundles for U-Net~\citep{unet}, UNet++~\citep{unetplusplus}, V-Net~\citep{vnet}, CMUNeXt~\citep{cmunext}, MedNeXt~\citep{mednext}, and UNETR~\citep{unetr}, covering both 2D and 3D segmentation tasks.

	\section{Training Transparency}
	\label{sec:transparency}
	
	A recurring frustration in medical image segmentation research is the opacity of the training process.
	Many frameworks report only a final score after training completes, leaving the researcher with little insight into how the model evolved, which cases are problematic, or whether training should be stopped early.
	MIPCandy treats training visibility as a first-class design goal: every training run automatically produces a rich set of artifacts that allow the researcher to monitor, diagnose, and communicate results without additional code.
	
	\paragraph{Console output and metric reporting.}
	During training, MIPCandy prints a structured summary after each epoch via the Rich library~\citep{rich}, including the current epoch, all tracked losses, validation scores, learning rate, epoch duration, and---when available---the estimated time of completion (ETC).
	After each validation pass, a per-case metric table is displayed, highlighting the worst-performing case so that the researcher can immediately identify failure modes.
	Appendix~\ref{app:console} (Figure~\ref{fig:console}) shows a representative console screenshot when resuming a previously interrupted training run, illustrating both the recovery mechanism and the per-epoch metric reporting.
	
\begin{figure}[t]
		\centering
		\begin{subfigure}[b]{0.48\columnwidth}
			\centering
			\includegraphics[width=\textwidth]{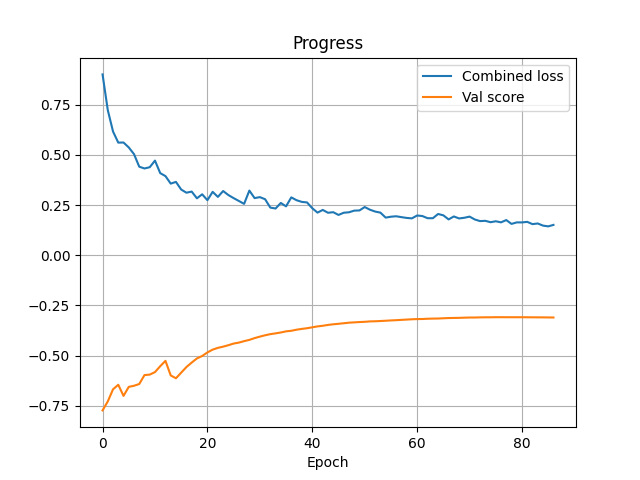}
			\caption{Combined loss and validation score.}
		\end{subfigure}
		\hfill
		\begin{subfigure}[b]{0.48\columnwidth}
			\centering
			\includegraphics[width=\textwidth]{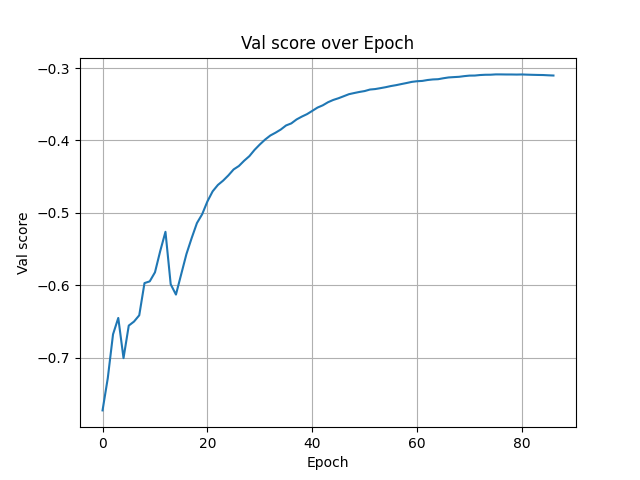}
			\caption{Validation score trajectory.}
		\end{subfigure}
		\caption{Training progress plots automatically generated by MIPCandy during a U-Net training run on the PH2 dermoscopy dataset. The validation score is the negated combined loss (Section~\ref{sec:training}); higher values indicate better performance.}
		\label{fig:progress}
	\end{figure}

	\paragraph{Training progress visualization.}
	At the end of each epoch, MIPCandy updates a set of metric curve plots saved to the experiment folder.
	These include combined loss and validation score on a single progress plot, as well as individual plots for each loss component (Dice, cross-entropy), per-class Dice scores, learning rate schedule, and epoch duration.
	Researchers can monitor these plots in real time via any file viewer or integrate them into slide decks and lab notebooks.
	Figure~\ref{fig:progress} shows the progress plot and validation score curve from a U-Net trained on PH2 for 90 epochs.
	
\begin{figure}[t]
		\centering
		\begin{subfigure}[b]{0.32\columnwidth}
			\centering
			\includegraphics[width=\textwidth]{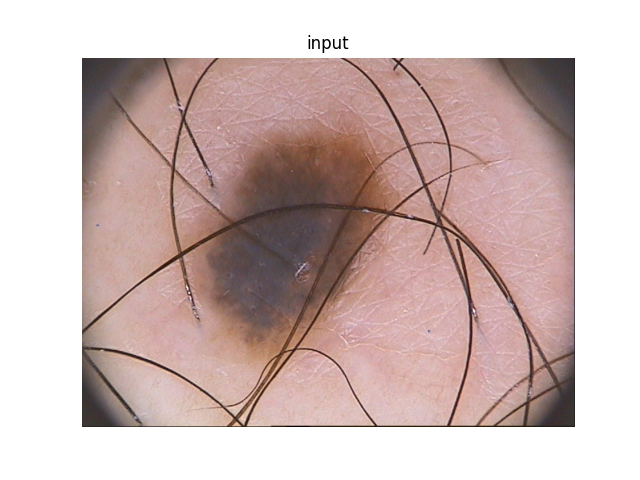}
			\caption{Input image.}
		\end{subfigure}
		\hfill
		\begin{subfigure}[b]{0.32\columnwidth}
			\centering
			\includegraphics[width=\textwidth]{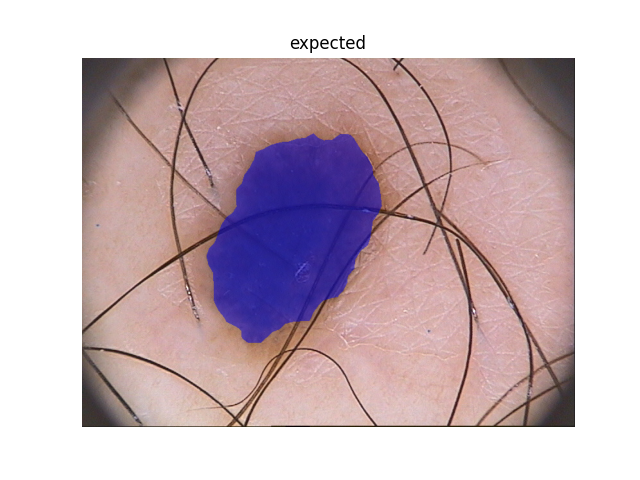}
			\caption{Expected (GT overlay).}
		\end{subfigure}
		\hfill
		\begin{subfigure}[b]{0.32\columnwidth}
			\centering
			\includegraphics[width=\textwidth]{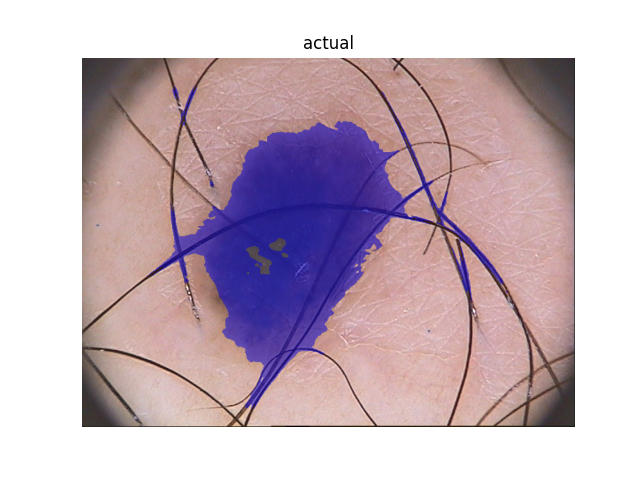}
			\caption{Actual (prediction overlay).}
		\end{subfigure}
		\caption{Worst-case prediction previews automatically saved during training. The framework identifies the validation case with the lowest score and generates overlays comparing the ground truth (b) and model prediction (c) against the input image (a). This example is from a U-Net trained on PH2.}
		\label{fig:preview}
	\end{figure}

\begin{figure}[t]
		\centering
		\begin{subfigure}[b]{0.48\columnwidth}
			\centering
			\includegraphics[width=\textwidth]{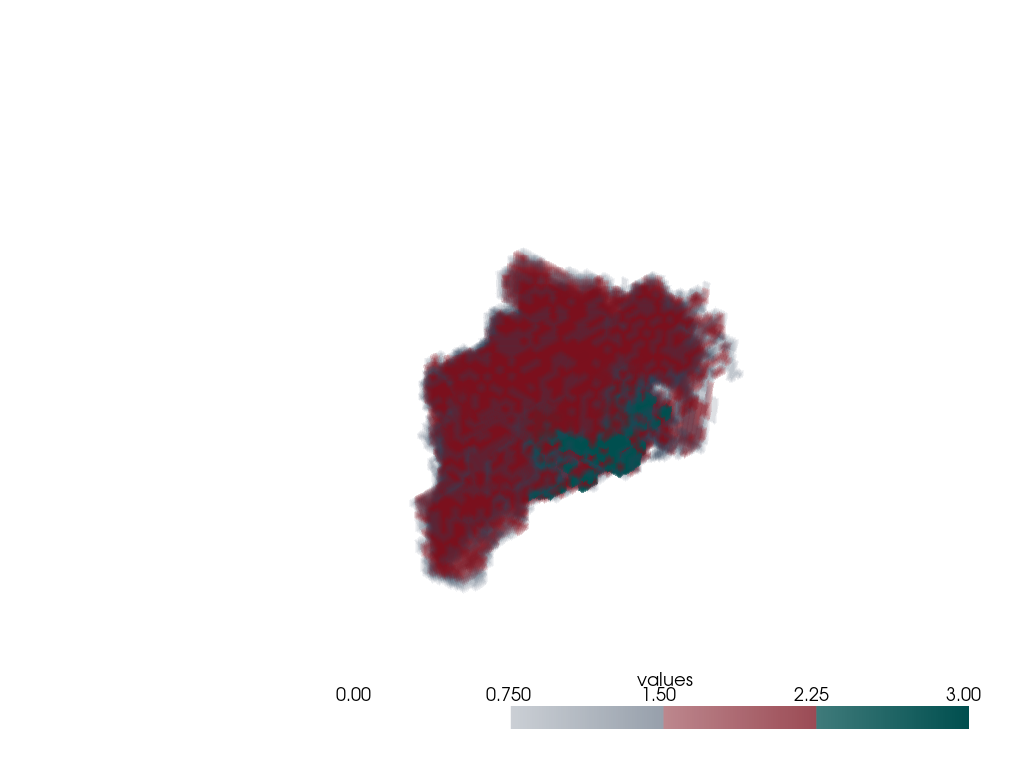}
			\caption{BraTS ground-truth label (4 classes).}
		\end{subfigure}
		\hfill
		\begin{subfigure}[b]{0.48\columnwidth}
			\centering
			\includegraphics[width=\textwidth]{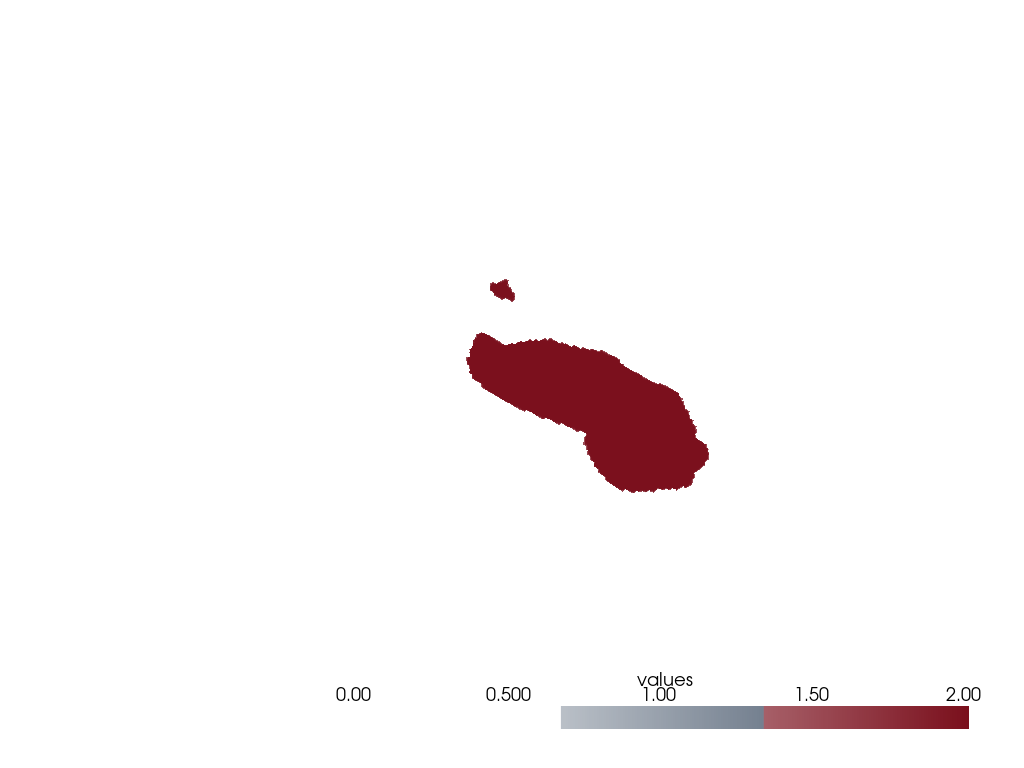}
			\caption{PANTHER~\citep{panther} predicted segmentation.}
		\end{subfigure}
		\caption{3D volume previews rendered via PyVista. MIPCandy automatically generates 3D visualizations of labels and predictions for volumetric segmentation tasks.}
		\label{fig:3d}
	\end{figure}

	\paragraph{Prediction previews and worst-case tracking.}
	After each validation epoch, the framework identifies the worst-performing validation case (by validation score) and saves a set of preview images: the raw input, the ground-truth label, the model's prediction, and two overlay composites---the expected overlay (ground truth superimposed on the input) and the actual overlay (prediction superimposed on the input).
	By always displaying the worst case rather than a random or cherry-picked example, this mechanism ensures that the researcher's attention is directed to the most informative failure mode.
	Figure~\ref{fig:preview} shows these previews from a 2D skin lesion segmentation experiment.
	For 3D volumes, \texttt{visualize3d()} renders the label and prediction as interactive PyVista~\citep{pyvista} meshes with automatic downsampling, as shown in Figure~\ref{fig:3d}.
	
	\paragraph{Validation score prediction.}
	After a configurable warm-up period (default: 20 epochs), MIPCandy fits a quotient regression model to the validation score trajectory and extrapolates the maximum achievable score and the epoch at which it will be reached.
	From these estimates the framework computes an ETC (Estimated Time of Completion) that is displayed after each validation epoch.
	This allows researchers to make informed decisions about early stopping, hyperparameter adjustment, or resource allocation without waiting for the full training run to complete.
	
	A full console screenshot illustrating both the recovery mechanism and the per-epoch output is provided in Appendix~\ref{app:console}.
	
\begin{figure}[t]
		\centering
		\includegraphics[width=0.75\columnwidth]{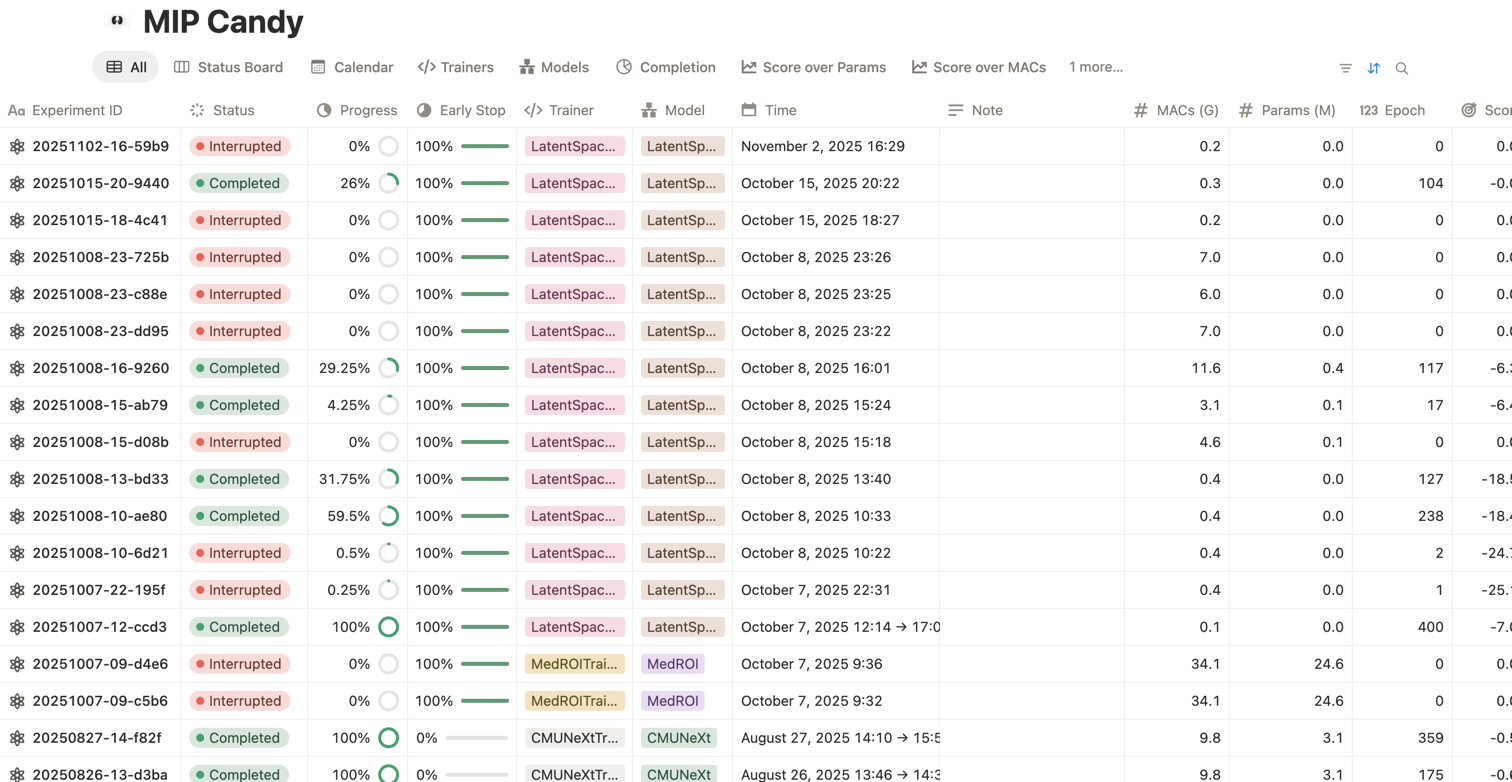}
		\caption{Notion frontend integration. MIPCandy automatically logs experiment metadata, progress, and scores to a Notion database.}
		\label{fig:notion}
	\end{figure}

	\paragraph{Frontend integrations.}
	For team-level experiment management, MIPCandy integrates with external tracking services via a lightweight \texttt{Frontend} protocol.
	Shipped frontends include Weights~\&~Biases~\citep{wandb}, Notion, and MLflow~\citep{mlflow}, and the \texttt{create\_hybrid\_frontend()} factory allows simultaneous logging to multiple services.
	Figure~\ref{fig:notion} shows a Notion database populated by MIPCandy, providing a persistent, shareable experiment ledger.
	
	\paragraph{Training state recovery.}
	Long-running 3D training jobs are frequently interrupted by hardware failures, preemption, or resource limits.
	MIPCandy serializes the full training state---optimizer, scheduler, criterion state dictionaries, and a state orb recording epoch, best score, and all training arguments---at every epoch.
	Training can be resumed via \texttt{recover\_from()} followed by \texttt{continue\_training()}, restoring the exact state and continuing from the interrupted epoch (Figure~\ref{fig:console}).

	\section{Case Studies}
	\label{sec:examples}

	This section demonstrates MIPCandy workflows on representative segmentation tasks, illustrating both the minimal code required and the artifacts produced by the framework.

	\subsection{2D Skin Lesion Segmentation}
	\label{sec:example-2d}

	The following script performs binary segmentation on the PH2 dermoscopy dataset~\citep{ph2} using a U-Net bundle.
	The complete pipeline---data loading, $k$-fold splitting, trainer configuration, and training---requires 8 lines of code:
	
	\begin{lstlisting}
		import torch
		from torch.utils.data import DataLoader
		from mipcandy.data import NNUNetDataset
		from mipcandy_bundles.unet import UNetTrainer
		
		device = "cuda" if torch.cuda.is_available() else "cpu"
		train, val = NNUNetDataset(folder="Dataset501_PH2", split="Tr").fold(fold=0)
		trainer = UNetTrainer(
		"experiments", DataLoader(train, batch_size=2, shuffle=True),
		DataLoader(val, batch_size=1), device=device,
		)
		trainer.num_classes = 1
		trainer.train(100)
	\end{lstlisting}
	
	Without a bundle, the same workflow requires implementing a single method on \texttt{SegmentationTrainer}:
	
	\begin{lstlisting}
		from typing import override
		from torch import nn
		from mipcandy.presets import SegmentationTrainer
		
		class MyTrainer(SegmentationTrainer):
		    @override
		    def build_network(self, example_shape: tuple[int, ...]) -> nn.Module:
		        from mipcandy_bundles.unet import make_unet2d
		        return make_unet2d(example_shape[0], self.num_classes)
	\end{lstlisting}
	
	Upon completion, the experiment folder contains model checkpoints, per-epoch metrics (CSV), all training curve plots shown in Section~\ref{sec:transparency}, and worst-case preview images.
	Evaluation on a held-out test set is equally concise:
	
	\begin{lstlisting}
		from mipcandy.evaluation import Evaluator
		from mipcandy.metrics import binary_dice
		from mipcandy_bundles.unet import UNetPredictor
		
		predictor = UNetPredictor("experiments/UNetTrainer/20240901-1234",
		example_shape=(3, 384, 384), device="cuda")
		evaluator = Evaluator(binary_dice)
		result = evaluator.predict_and_evaluate("test_images/", "test_labels/", predictor)
		print(result.mean_metrics)
	\end{lstlisting}

	\subsection{3D Volumetric Segmentation}
	\label{sec:example-3d}

	For 3D tasks, MIPCandy's dataset inspection system automates the determination of patch shapes and foreground sampling rates.
	The following script trains a multiclass 3D segmentation model on the BraTS 2021 brain tumor dataset~\citep{brats} with deep supervision and ROI-based patch sampling:
	
	\begin{lstlisting}
		import torch
		from torch.utils.data import DataLoader
		from mipcandy.data import NNUNetDataset
		from mipcandy.data.inspection import inspect, RandomROIDataset
		from mipcandy_bundles.unet import UNetTrainer
		
		device = "cuda" if torch.cuda.is_available() else "cpu"
		dataset = NNUNetDataset(folder="Dataset320_BRaTS", split="Tr")
		train_full, val_full = dataset.fold(fold=0)
		
		# Inspect dataset to compute ROI shape and class distribution
		annotations = inspect(train_full)
		train = RandomROIDataset(annotations, batch_size=2)
		val = RandomROIDataset(
		inspect(val_full), batch_size=1, oversample_rate=0
		)
		
		trainer = UNetTrainer(
		"experiments", DataLoader(train, batch_size=2, shuffle=True),
		DataLoader(val, batch_size=1), device=device,
		)
		trainer.num_dims = 3
		trainer.num_classes = 4
		trainer.deep_supervision = True
		trainer.train(200, early_stop_tolerance=20)
	\end{lstlisting}
	
	The \texttt{inspect()} call scans the training set to compute per-case foreground bounding boxes, class distributions, and intensity statistics.
	\texttt{RandomROIDataset} uses these annotations to sample patches of a statistically determined shape, with 33\% of patches forced to contain foreground voxels.
	Deep supervision is enabled by setting a single flag; the trainer automatically wraps the loss function, computes scale-dependent weights, and generates multi-resolution targets.
	The framework produces 3D preview renderings (Figure~\ref{fig:3d}), per-class Dice curves, and all other transparency artifacts described in Section~\ref{sec:transparency}.

	\section{Conclusion}
	\label{sec:conclusion}

	We have presented MIPCandy, a modular, PyTorch-native framework for medical image segmentation that prioritizes four qualities: \emph{flexibility} in swapping components, \emph{transparency} during training, \emph{usability} through minimal-code setup, and \emph{extensibility} via a bundle ecosystem.
	
	The framework provides a complete pipeline from data loading through training and inference to evaluation.
	A functional segmentation workflow can be obtained by implementing a single method---\texttt{build\_network}---while all infrastructure (loss selection, optimizer configuration, checkpointing, metric tracking, deep supervision, EMA, validation score prediction, and experiment tracking) is handled by the framework with researched defaults.
	At the same time, every component is independently usable and replaceable, allowing incremental adoption into existing PyTorch codebases.
	
	The key technical contributions---\texttt{LayerT} for compositional module configuration, built-in training transparency with worst-case tracking and validation score prediction, dataset inspection with ROI-based patch sampling, and the bundle ecosystem---address practical pain points in medical image segmentation research.
	Unlike fully automated pipelines that treat training as a black box, MIPCandy ensures that the researcher retains full visibility into and control over every stage of the process.
	
	MIPCandy is open-source under the Apache-2.0 license and is actively developed.
	Future work includes expanding the metric library with surface-distance metrics (Hausdorff distance, average symmetric surface distance), adding sliding window inference for large volumes, supporting semi-supervised and self-supervised learning paradigms, and extending the bundle ecosystem with task-specific bundles for detection and registration.

	\bibliographystyle{unsrtnat}
	\bibliography{references}

	\appendix

	\section{Console Output}
	\label{app:console}

\begin{figure*}[h]
		\centering
		\includegraphics[width=\textwidth]{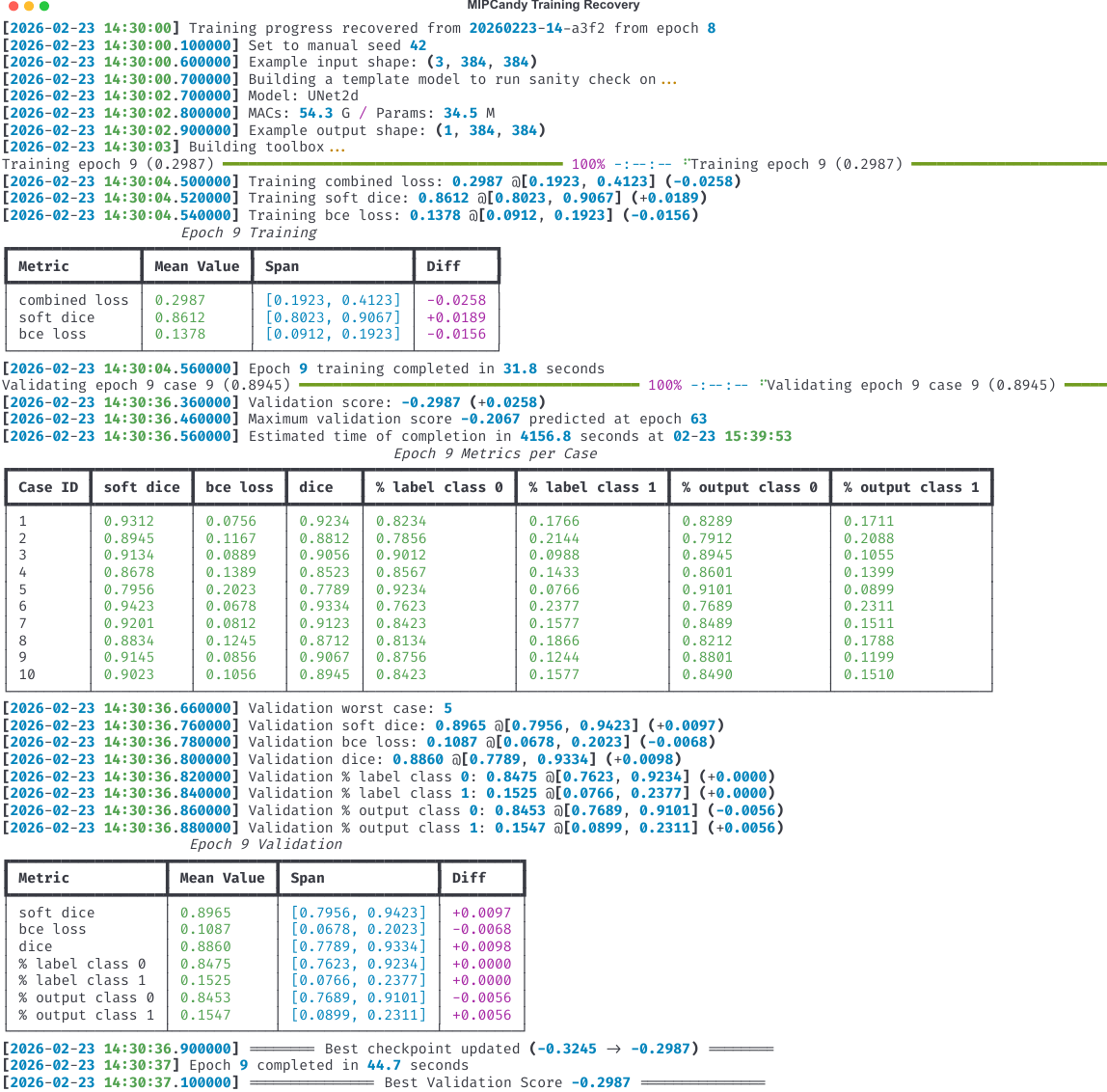}
		\caption{Console interface during training state recovery. The output shows a single epoch after recovery: sanity check, training metrics with a structured summary table, per-case validation metrics with per-class statistics, score prediction with ETC, and checkpoint management.}
		\label{fig:console}
	\end{figure*}

	Figure~\ref{fig:console} shows the full console output when resuming a previously interrupted training run via \texttt{recover\_from()} and \texttt{continue\_training()}.
	The framework restores the optimizer, scheduler, and training tracker from a previous checkpoint and resumes from the interrupted epoch.
	Each epoch produces a structured metric summary, a per-case validation table with per-class label and prediction statistics, and aggregated validation metrics.
	The worst-performing validation case is highlighted, and the estimated time of completion (ETC) is displayed based on quotient regression of the validation trajectory.

\end{document}